# Battle royale optimizer with a new movement strategy

Sara Akan[1][0000-0001-7822-1549] and Taymaz Akan[2][0000-0003-4070-1058]

[1] Department of Computer programing, Ayvansaray University, Istanbul, Turkey
[2] Department of Software Engineering, Ayvansaray University, Istanbul, Turkey
[3] Faculty of Electrical Eng. & Informatics, University of Pardubice, Pardubice, Czech Republic
{saraakan, taymazakan}@ayvansaray.edu.tr

**Abstract.** Gamed-based is a new stochastic metaheuristics optimization category that is inspired by traditional or digital game genres. Unlike SI-based algorithms, individuals do not work together with the goal of defeating other individuals and winning the game. Battle royale optimizer (BRO) is a Gamed-based metaheuristic optimization algorithm that has been recently proposed for the task of continuous problems. This paper proposes a modified BRO (M-BRO) in order to improve balance between exploration and exploitation. For this matter, an additional movement operator has been used in the movement strategy. Moreover, no extra parameters are required for the proposed approach. Furthermore, the complexity of this modified algorithm is the same as the original one. Experiments are performed on a set of 19 (unimodal and multimodal) benchmark functions (CEC 2010). The proposed method has been compared with the original BRO alongside six well-known/recently proposed optimization algorithms. The results show that BRO with additional movement operator performs well to solve complex numerical optimization problems compared to the original BRO and other competitors.

## 1 Introduction

Metaheuristic is a multipurpose algorithm that solves problems in an iterative manner, trying to find the best feasible option among a number of applicant solutions with a quality measure [1]. Metaheuristic optimization methods have attracted the attention of scientists and engineers from a number of disciplines, as they serve a major role both in scientific and industrial use in many practical challenges [2]. Over the last three decades, various optimization techniques have been proposed on various sources of inspiration. Battle royale optimization (BRO) [3] is a recently proposed optimization algorithm that is motivated by a digital games genre called the "royal battle". BRO is a population-based optimization algorithm that simulates the deathmatch play mode of Player Unknown's Battlegrounds (PUBG) [4]. BRO aims at solving single-objective optimization in continuous problem spaces [5]. Although several optimization algorithms have been introduced in recent years, the BRO algorithm is of particular im-portance among these algorithms. This is because this algorithm has opened a new horizon for metaheuristic algorithms. Before this algorithm was developed, metaheuristic algorithms consisted of three main categories: Evolutionary algorithms (EA); Swarm Intelligence (SI); and physical phenomena algorithms. But BRO introduced a new metaheuristic category called game-based.

EA-based algorithms are inspired by Darwin's theory of biological Evolution, mimicking concepts of reproduction, mutation, and selection to achieve the best possible solution. Genetic Algorithm (GA) [6], Evolution Strategies (ES) [7], Tabu Search (TS) [8], Simulated-Annealing (SA) [9], Differential Evolution (DE) [10],Biogeography-Based Optimizer (BBO) [11], Forest Optimization Algorithm (FOA) [12], etc. can be mentioned as the most well-known EA paradigms.

SI-based algorithms mimic the collective behavior of various kinds of the living being, such as humans, animals, insects, single-cells, etc., in the population. Although not every individual is intelligent alone, their collective behavior leads to an intelligent search scheme. Some of well-known and recently developed SI-based algorithms are Particle Swarm Optimization [13], Ant colony optimization [14], Cat swarm optimization [15], Artificial Bee Colony(ABC)[16], Cuckoo Search(CS) [17], Firefly Algorithm (FA) [18], Animal migration optimization [19], Dolphin echolocation (DE) [20], Chimp optimization algorithm (COA) [21], Human mental search (HMS) [22], and Selfish Herd Optimizer (SHO) [23], Group Search Optimizer (GSO) [24] as well as others.

To reach the optimal solution physics-based algorithms make use of the laws of physics such as gravitational force, inertia force, electromagnetic force, etc. The movement and communication between populations are conducted by means of these rules [25]. Examples of physics-based methods are Quantum stochastic optimization (QSO) [26], Central Force Optimization (CFO) [27], Gravitational Search Algorithm (GSA)[28], Charged System Search (CSS) [29], Black Hole (BH) [30], Optics inspired optimization (OIO) [31], Water Evaporation Optimization (WEO) [32],Yin-Yang-Pair Optimization (YYPO)[33], Thermal exchange optimization [34], Electromagnetic Field Optimization (EFO) [35], Billiards-inspired optimization algorithm (BOA) [36], and others.

Exploration and exploitation are two critical factors in the success of an optimization algorithm. Every optimization algorithms have their own strategy for performing exploration and exploitation mechanism [28]. In order to deliver a successful outcome in an optimization algorithm, the balance between exploration and exploitation must be maintained. Exploration aims to escape from local optima by exploring among new solution-candidates in the unvisited areas of problem space. On the other hand, exploitation focuses on the area where exploitation focuses on the best solution to search more accurately. E.g., in PSO $p_{best}$ provides exploitation alongside $c_1$ coefficient parameter while $g_{best}$ provides exploration alongside $c_2$ coefficient parameter. Moreover, in some cases, an additional local search step is applied to enhance exploitation [37].

According to the famous mathematician Leonhard Euler, everything in nature is ultimately either minimized or maximized. Whether plants or animals, living beings have been able to adapt to the environment over time, and they found a way to survive. Nature has always seen and will continue to witness far-reaching changes; as a result, nature can be exposed to new changes and problems that must find new solutions to adapt to them. In fact, the scientific and engineering community is not unlike nature: every day, new issues may arise that at times require a new method of solution. Hence, it makes sense to be inspired by nature to develop optimization algorithms to minimize or maximize the problems. This claim could be a confirmation of the No Free Lunch (NFL) theorem [38]. NFL is a response to critics who claim that there are many optimization algorithms and that there is no need for new algorithms. The NFL proves that an optimization algorithm may not outperform others in meeting all optimization challenges. There is no heuristic algorithm yet that solves all optimization problems by providing superior performance. Therefore, it is necessary to propose new approaches as well as to improve existing ones.

BRO is a recently-proposed population-based metaheuristic optimization algorithm for continuous optimization problems that is inspired by strategies of multiplayer video game genre (Battle royale games). There are various kinds of Battle Royale games, the most important of which are Player Unknown's Battlegrounds (PUBG) [4], Call of Duty: Warzone [39], Apex Legends [40], Counter-Strike: Global Offensive [41], and Ring of Elysium [42]. However, these games are not very different from each other in game strategy. In fact, BRO mimics the solo-deathmatch mode of PUBG.

Like other stochastic algorithms, search agents randomly distributed over the problem space using a uniform distribution. BRO generates randomly $N$ solutions using Eq. 1.

$$x_{i,d} = r\left(ub_d - lb_d\right) + lb_d \tag{1}$$

where $i = 1 \cdots N$, $d = 1 \cdots D$. $N$ is the number of possible solutions in the population. $D$ is the dimension of the problem space. $lb_d$ and $ub_d$ are the lower and the upper bounds of $d^{th}$ dimension, respectively. In the following, the fitness value of each solution is compared to the fitness value of the nearest neighbor with regard to the Euclidean distance. And then, the solution with the better fitness value will be declared the winner and the other the loser. Next, the loser one will be damaged, and the damage count of it will be increased by one. On the contrary, the damage of the winner will be reset to zero. The operator is expressed as: $x_{dam}.damage = x_{dam}.damage + 1$ and $x_{vic}.damage = 0$, where $dam$ and $vic$ are the indexes of the loser and winner solutions, respectively. Meanwhile, loser one attempts to move toward the best solution to locate the better position. The new position vector of the loser solution can be calculated by Eq. 2.

$$x_{dam,d} = x_{dam,d} + r(x_{best,d} - x_{dam,d}), \tag{2}$$

where $r$ is a randomly-generated number from a uniform distribution [0,1] and $x_{dam,d}$ is the position of the damaged solution in $d^{th}$ dimension. Moreover, If a solution has a worse fitness value than his neighbor through a pre-determined number of times in a row, it will not be a solution candidate. For this purpose, the solution will be eliminated from the set of the solutions and will be replaced by a random one according to Eq. 1. It is worth noting that to provide exploitation, the problem space shrinks down towards the best solution in every $\Delta$ iteration. The initial value for $\Delta$ is calculated per $\Delta = MaxCicle/log_{10}(MaxCicle)$. Then, it will be updated through $\Delta = \Delta + round(\frac{\Delta}{2}))$ . $MaxCicle$ is the maximum number of iteration herein. To put this concept lower and upper bound of each problem dimension will shrink towards the best solution as follows:

$$lb_d = x_{best,d} - SD\left(\overline{x_d}\right)$$
$$ub_d = x_{best,d} + SD\left(\overline{x_d}\right) \tag{3}$$

where $SD(\overline{x_d})$ indicates the standard deviation of all solutions of the population in $d^{th}$ dimension and $x_{best,d}$ refers to the position of the best solution ever found.

## 2    A new movement strategy

The new movement strategy is characterized by keeping the individual motion and the expansion of the search area. In doing so, it can reflect changes in individual speed. In this case, by using an additional parameter, the individual in motion updates and maintains another parameter to use in the next iteration. This parameter is randomly generated at first and is later updated. Subsequently, its value would be closely associated with an individual search capability. The greater the value of the movement operator, the greater the individual speed will be. To do a global search, the individuals would have a larger step. Their global capacity for optimization is strong, and their capacity for local search is weak. To do a global search, the individuals would have a larger step. Their global capacity for optimization is strong, and their capacity for local search is weak. The smaller the value of the movement operator, the smaller the size of the individual step will be. If the movement step is too large, an algorithm can fail to achieve or converge to the optimal solution, resulting in the algorithm not converging.

For the moving an individual towards the best one the Eq. 2 has been reformulated as follows:

$$\lambda_{dam,d} = r_1 x_{best,d} + r_2(\lambda_{dam,d} - x_{dam,d}),$$
$$x_{dam,d} = \lambda_{dam,d} + x_{dam,d} \tag{4}$$

where $r_1$ and $r_2$ are two randomly-generated numbers from a uniform distribution [0,1] and $\lambda_{dam,d}$ is also a combination randomly-generated numbers from a uniform distribution [0,1] that will be updated through the iterations. This modified moving strategy provides more randomness in both exploration and exploitation.

**Table 1**. Unimodal benchmark test functions

| Function | Name | Range | Shift position |
|---|---|---|---|
| $f_1(\mathbf{x}) = \sum_{i=1}^{n} x_i^2$ | Sphere | [-100,100] | [-30, -30,···, -30] |
| $f_2(\mathbf{x}) = \sum_{i=1}^{n} |x_i| + \prod_{i=1}^{n} |x_i|$ | Schwefel 2.20 | [-10,10] | [-3, -3,···, -3] |
| $f_3(\mathbf{x}) = \sum_{i=1}^{n} \left( \sum_{j=1}^{i} x_j \right)^2$ | Rotated hyper-ellipsoids | [-100,100] | [-30, -30,···, -30] |
| $f_4(\mathbf{x}) = \max_{i=1,...,n} |x_i|$ | Schwefel 2.21 | [-100,100] | [-30, -30,···, -30] |
| $f_5(\mathbf{x}) = \sum_{i=1}^{n-1} \left[ 100(x_{i+1} - x_i^2)^2 + (x_i - 1)^2 \right]$ | Rosenbrock | [-30,30] | [-15, -15,···, -15] |
| $f_6(\mathbf{x}) = \sum_{i=1}^{n} ([x_i + 00.5])^2$ | Step | [-100,100] | [-750, -750,···, -750] |
| $f_7(\mathbf{x}) = \sum_{i=1}^{n} i x_i^4 + rand[0,1)$ | Quartic | [-128,128] | [-25, -25,···, -25] |

## 3   Experimental results and performance evaluation

A comprehensive experimental evaluation and comparison were performed to evaluate the performance of the BRO with a random inertia weight. Like the original BRO well-known PSO (Particle Swarm Optimization) algorithm and five recent proposed optimization algorithms: ALO (The Ant Lion Optimizer) [43], MFO (Moth-flame optimization algorithm) [44], MVO (Multi-Verse Optimizer) [45], SCA (A Sine Cosine Algorithm) [46], and WOA (The Whale Optimization Algorithm) [47] have been selected as competitors. Moreover, the original BRO has also been used in comparisons. The parameters for the control of all algorithms have been set in the initial papers according to recommendations. In addition, all algorithms were carried out and implemented in MATLAB and Conducted on a Core i7-7700 HQ 2.81 Processor with 32 GB of RAM. This paper also uses 19 well-known benchmark functions for comparison that include unimodal and multi-modal properties, like the original study. All these functions have been listed in Tables 1–3. For every algorithm, the maximum number of iteration and population size is 500 and 100, respectively. Also, the threshold value for maximum damage is 3, and both $r_1$ and $r_2$ are randomly generated number uniformly distributed in the range [0,1]. All of the following test results are achieved by taking an average of 25 independent runs each. In addition, the median and standard variation of the fitness evaluation values are used as performance measures in 25 independent runs. Tables 4, 5, and 6 for all unimodal and multimodal test functions, have provided the experimental results.

Table2. Multimodal benchmark test functions

| Function | Name | Range | Shift position |
|---|---|---|---|
| $f_8(\mathbf{x}) = \sum_{i=1}^{n} x_i sin(\sqrt{|x_i|})$ | Schwefel | [-500,500] | [-300,⋯, -300] |
| $f_9(\mathbf{x}) = 10n + \sum_{i=1}^{n}(x_i^2 - 10cos(2\pi x_i))$ | Rstrigin | [-5.12,5.12] | [-2,⋯, -2] |
| $f_{10}(\mathbf{x}) = -20.exp(-0,2\sqrt{\frac{1}{n}\sum_{i=1}^{n} x_i^2})$ $-exp(\frac{1}{n}\sum_{i=1}^{n} cos(2\pi x_i)) + 20 + exp(1)$ | Ackley | [-32,32] | |
| $f_{11}(\mathbf{x}) = 1 + \sum_{i=1}^{n} \frac{x_i^2}{4000} - \prod_{i=1}^{n} cos(\frac{x_i}{\sqrt{i}})$ | Griewank | [-600,600] | [-400,⋯, -400] |
| $f_{12}(\mathbf{x}) = \frac{\pi}{n} \times \{10sin^2(\pi y_1) + \sum_{i=1}^{n-1}(y_i - 1)^2$ $[1 + 10sin^2(\pi y_{i+1})] + (y_n - 1)^2\} +$ $\sum_{i=1}^{n} u(x_i, 10, 100, 4)$ $y_i = 1 + \frac{1}{4}(x_i + 1)$ $u(x_i, a, k, m) = \begin{cases} k(x_i - a)^m & \text{if } x_i > a \\ 0 & \text{if } -a \leq x_i \leq a \\ k(-x_i - a)^m & \text{if } x_i < -a \end{cases}$ | penalized | [-50,50] | [-30,⋯, -30] |
| $f_{13}(\mathbf{x}) = 0.1\{sin^2(3\pi x)$ $+(x_i - 1)^2(1 + sin^2(3\pi y))$ $+(x_n - 1)^2(1 + sin^2(2\pi x_n))\}$ $+\sum_{i=1}^{n} u(x_i, 5, 100, 4)$ | Levi | [-50,50] | [-100,100] |

**Table3**. Multimodal benchmark test functions with fixed- dimension

| Function | Dim | Range |
|---|---|---|
| $f_{14} = \left(\frac{1}{500} + \sum_{j=1}^{25}\frac{1}{j + \sum_{i=1}^{2}(x_i - a_{ij})^6}\right)^{-1}$ | 2 | [-65,65] |
| $f_{15} = \sum_{i=1}^{11}\left[a_i - \frac{x_1(b_i^2 + b_i x_2)}{b_i^2 + b_i x_3 + x_4}\right]^2$ | 4 | [-5,5] |
| $f_{16} = 4x_1^2 - 2.1x_1^4 + \frac{1}{3}x_1^6 + x_1 x_2 + 4x_2^2 + 4x_2^4$ | 2 | [-5,5] |
| $f_{17} = \left(x_2 - \frac{5.1}{4\pi^2}x_1^2 + \frac{5}{\pi}x_1 - 6\right)^2 + 10\left(1 - \frac{1}{8\pi}\right)\cos(x_1) + 10$ | 2 | [-5,5] |
| $f_{18} = [1 + (x_1 + x_2 + 1)^2(19 - 14x_1 + 3x_1^2 - 14x_2 + 6x_1 x_2 + 3x_2^2)]$ $\times [30 + (2x_1 - 3x_2)^2$ $\times (18 - 32x_1 + 12x_1^2 + 48x_2 + 36x_1 x_2 + 27x_2^2)]$ | 2 | [-2,2] |
| $f_{19} = -\sum_{i=1}^{4} c_i e^{\left(-\sum_{i=1}^{3} a_{ij}(x_j - p_{ij})^2\right)}$ | 3 | [1,3] |

**Table 4** Results for unimodal benchmark test functions of 30 dimension

| F | | M-BRO | BRO | PSO | ALO | MFO | MVO | SCA | GOA |
|---|---|---|---|---|---|---|---|---|---|
| F1 | mean | **1.2012e-26** | 3.0353e-09 | 8.131e-08 | 1.e-06 | 0.937433 | 0.24190 | 0.139829 | 1028.242 |
| | Std | **1.6456e-26** | 4.1348e-09 | 1.567e-07 | 1.e-06 | 0.547602 | 0.04815 | 0.213883 | 2921.282 |
| | time | 3.598942 | 3.536958 | 1.1516 | 76.82223 | 0.5428634 | 0.7795 | **0.213883** | 1264.051 |
| F2 | mean | **1.3350e-15** | 0.000046 | 0.1282 | 20.52849 | 28.501319 | 0.29813 | 0.000979 | 77.97797 |
| | Std | **9.8358e-16** | 0.000024 | 0.2889 | 31.56666 | 19.846522 | 0.07011 | 0.001209 | 48.421113 |
| | time | 3.290291 | 3.577353 | 1.1460 | 77.20241 | **0.5510458** | 0,71363 | 0. 5723780 | 1332.949 |
| F3 | mean | **4.2811e-08** | 54.865255 | 103.5336 | 127.4139 | 12432.852 | 13.2258 | 1524.3665 | 5095.518 |
| | Std | **6.2916e-08** | 16.117329 | 77.8071 | 43.86708 | 8927.4874 | 5.38360 | 1339.0129 | 5875.806 |
| | time | 3.829758 | 3.710890 | 1.1226 | 77.78005 | 1.449506 | 1.62891 | **1.436205** | 1206.811 |
| F4 | mean | **7.7175e-07** | 0.518757 | 3.0470 | 8.659458 | 30.626058 | 0.52880 | 16.26705 | 1.775508 |
| | Std | **1.4126e-06** | 0.403657 | 1.9104 | 3.145536 | 10.680434 | 0.22757 | 9.442185 | 0.989782 |
| | time | 3.834288 | 3.532575 | 1.1399 | 77.5049 | **0.5653759** | 0,74207 | 0.5887966 | 1303.754 |
| F5 | mean | **27.03186** | 99.93684 | 59.0941 | 70.59842 | 1128.7042 | 329.447 | 782.05773 | 754.8616 |
| | Std | **0.4024602** | 82.862358 | 44.7770 | 71.72374 | 1327.9809 | 566.273 | 2102.6807 | 2748.332 |
| | time | 5.096510 | 3.948993 | 1.2316 | 77.04758 | **0.6677657** | 0. 8423 | 0. 720768 | 423.1110 |
| F6 | mean | 2.257257 | **2.8731e-08** | 9.332e-08 | 1.e-06 | 1601.2818 | 0.23379 | 4.097225 | 0.000003 |
| | Std | 0.3160390 | **1.8423e-08** | 2.255e-07 | 1.e-06 | 3741.6075 | 0.05323 | 0.599685 | 0.000002 |
| | time | 4.221259e | 3.551993 | 1.1685 | 77.13036 | 0.5411606 | 0.74088 | 5.861539 | 424.5775 |
| F7 | mean | **7.8361e-04** | 0.000368 | 0.0215 | 0.023584 | 2.211140 | 0.00798 | 0.021428 | 4.330057 |
| | Std | **5.8031e-04** | 0.000094 | 0.0097 | 0.00849 | 3.461601 | 0.00327 | 0.01812 | 9.45593 |
| | time | 5.196062e | 4.307858 | 1.5415 | 77.5254 | **1.082169** | 1.26860 | 1.102313 | 1242.990 |

Table 4 proves that the modified BRO provides the best results in six out of seven unimodal functions with respect to mean and std. It is clear that there is a great improvement in the results. As it is clear, GOA performs worst among the others. Also,

Table 5 shows that in five of seven multimodal functions, BRO outperforms others in terms of mean and outperforms five out of seven respect to std. As is evident from these tables, there is a great improvement in results. Such that, for F1, the mean value of the proposed method is 1.20128e-26, while for the original BRO is 3.0353e-09. Coming to the fixed-dimension multimodal functions, the modified BRO has not performed better in just one case (F14). In the rest of the function, the proposed method provides competitive results.

**Table 5** Results for multimodal benchmark test functions of 30 dimension

| F | | M-BRO | BRO | PSO | ALO | MFO | MVO | SCA | GOA |
|---|---|---|---|---|---|---|---|---|---|
| F8 | mean | **-4.5130e+03** | -7035.210 | **-9.39e+31** | -5601.864 | -9439.105 | -7872.021 | -4118.072 | -6759.6244 |
| | Std | 5.3803e+02 | 712.3326 | 4.22e+32 | 362.17167 | 933.4169 | 651.30106 | **295.8680** | 899.29158 |
| | time | 3.871863 | 3.782519 | 1.1632 | 77.28404 | **0.705079** | 0.7452922 | 0.732390 | 1259.423 |
| F9 | mean | **0** | 48.27535 | 54.2471 | 60.851573 | 107.59104 | 108.04483 | 17.584379 | 174.76975 |
| | Std | **0** | 14.09458 | **13.4287** | 22.120831 | 24.913857 | 32.191667 | 20.95633 | 29.134588 |
| | time | 3.811548 | 3.771521 | 1.2822 | 77.06642 | **0.621141** | 0.8954284 | 0.635833 | 1270,283 |
| F10 | mean | **3.3573e-14** | 0.350724 | 0.9906 | 2.053675 | 6.363405 | 0.712974 | 10.13371 | 3.366269 |
| | Std | **3.595e-14** | 0.688712 | 0.8038 | 0.796278 | 8.534955 | 0.693826 | 9.537759 | 4.154672 |
| | time | 3.646655 | 3.784324 | 1.2443 | 77.20121 | **0.647817** | 0.9157972 | 0.684157 | 1209.365 |
| F11 | mean | **0** | 0.001373 | 0.0089 | 0.010494 | 0.751211 | 0.479614 | 0.455533 | 7.231566 |
| | Std | **0** | 0.010796 | 0.0113 | 0.016548 | 0.179238 | 0.159060 | 0.275454 | 23.778806 |
| | time | 3.84395 | 3.844791 | 1.2318 | 77.07877 | 0.735585 | 1.028005 | 0.775249 | 1213.085 |
| F12 | mean | **0.130451** | 0.369497 | 30.4105 | 6.933983 | 1.734171 | 0.553024 | 1.184349 | 4323306.37 |
| | Std | **2.8956e-02** | 0.601450 | 9.4299 | 3.807565 | 1.328050 | 0.624414 | 1.270975 | 1616521.88 |
| | time | 5.694005 | 5.278253 | 1.3266 | 78.24491 | **1.965486** | 2.268537 | 2.036552 | 1214.762 |
| F13 | mean | 0.628109 | **0.000004** | 41.1520 | 0.006889 | 2.047761 | 0.032335 | 19.14310 | 18404721.8 |
| | Std | 0.110578 | **0.000020** | 9.6473 | 0.009962 | 1.909983 | 0.013044 | 71.92344 | 81308523.2 |
| | time | 3.304954 | 3.303456 | 1.3337 | 78.24868 | **1.986649** | 2.291993 | 2.154423 | 1191.996 |

**Table 6** Results for fixed-dimension multimodal benchmark test functions

| F | | M-BRO | BRO | PSO | ALO | MFO | MVO | SCA | GOA |
|---|---|---|---|---|---|---|---|---|---|
| F14 | mean | 0.998004 | **0.9980** | **0.9980** | 1.157048 | 0.998004 | 0.998004 | 1.077373 | 1.791015 |
| | Std | 0.6285483 | **0** | **0** | 0.371931 | **0** | 4.4661e-12 | 0.396819 | 1.144130 |
| | time | 3.565870 | 3.60161 | 2.7190 | 8.726093 | **0.282815** | 2.960769 | 0.289890 | 824.6873 |
| F15 | mean | **04.5511e-04** | 0.00047 | 0.0005105 | 0.001544 | 0.000748 | 0.000624 | 0.000845 | 0.007727 |
| | Std | 3.00125e-04 | **0.00026** | 0003.3323 | 0.003925 | 0.000270 | 0.000315 | 0.000398 | 0.008465 |
| | time | 2.486470 | 2.36244 | 0.9858 | 1.140260 | 0.277466 | 0.3253819 | **0.272637** | 158.6677 |
| F16 | mean | **-1.0316** | **-1.0316** | **-1.0316** | -1.03162 | -1.03162 | -1.031628 | -1.031619 | -1.03162 |
| | Std | 8.1149e-06 | **5.995e-16** | 6.798e-16 | 6.872e-16 | 6.79e-16 | 4.7670e-08 | 0.000007 | 7.194e-16 |
| | time | 2.015800 | 1.950645 | 0.8377 | 6.266332 | 0.25199 | 0.2871051 | **0.242102** | 79.82012 |
| F17 | mean | **0.397887** | **0.397887** | 0.3979 | **0.397887** | 0.39788 | 0.397887 | 0.398227 | 0.410236 |
| | Std | **0** | **0** | **0** | **0** | 7.42e-16 | **0** | 0.000260 | 0.064367 |
| | time | 1.959311 | 1.975396 | 0.7929 | 5.986799 | 0.21526 | 0.2524479 | **0.206958** | 117.7970 |
| F18 | mean | **3.000000** | **3.000000** | **3.000000** | **3.000000** | **3.000000** | 3.000001 | 3.000002 | **3.000000** |
| | Std | **2.5895e-16** | **2.5895e-16** | 1.344e-15 | 9.012e-14 | **1.88e-15** | 3.3333e-07 | 0.000003 | 0,000101 |
| | time | 1.979300 | 2.104709 | 0.7916 | 6.17052 | 0.20697 | 0.2461370 | **0.192042** | 84,81287 |
| F19 | mean | **-3.86208** | -3.86278 | -3.8628 | -3.86278 | -3.8627 | -3.862782 | -3.856386 | -3.8566 |
| | Std | 5.4290e-04 | **2.006e-15** | 2.266e-15 | 6.225e-15 | 2.266e-15 | 8.0539e-08 | 0.003206 | 0.0032 |
| | time | 2.884435 | 4.461555 | 1.0814 | 9.011699 | 0.302513 | 0.3353661 | **0.299619** | 167.7970 |

## 4    Conclusion

In this article, a modified BRO with a new movement operator has been proposed to provide a better balance between exploration and exploitation. Within the modified BRO, there are no user-set parameters, nor are kinds of other user-intervention required. At the same time, this algorithm maintains the complexity of the original one. Experiments were performed on 19 well-known (unimodal and multimodal) benchmark functions (CEC 2010). To verify the performance of the modified BRO, the original BRO, besides the well-known PSO algorithm and five recent proposed optimization algorithms ALO, MFO, MVO, SCA, and GOA, were also tested for comparison. Results indicate that the BRO with random inertia weight performs well to address complex numerical optimization problems, compared to the original BRO and other competitors.

**Conflict of interest** Authors declare that they have no conflict of interest.

## References

1. Yu JJQ, Li VOK, Lam AYS Optimal V2G scheduling of electric vehicles and Unit Commitment using Chemical Reaction Optimization. In: 2013 IEEE Congress on Evolutionary Computation, 20-23 June 2013 2013. pp 392-399. doi:10.1109/CEC.2013.6557596
2. Lazar A (2002) Heuristic knowledge discovery for archaeological data using genetic algorithms and rough sets. In:  Heuristic and Optimization for Knowledge Discovery. IGI Global, pp 263-278
3. Rahkar Farshi T (2020) Battle royale optimization algorithm. Neural Computing and Applications. doi:10.1007/s00521-020-05004-4
4. contributors W (2020) PlayerUnknown's Battlegrounds — Wikipedia, The Free Encyclopedia.
5. Agahian S, Akan T (2021) Battle royale optimizer for training multi-layer perceptron. Evolving Systems:1-13
6. Holland J (1975) Adaptation in natural and artificial systems: an introductory analysis with application to biology. Control and artificial intelligence
7. Schwefel H-P (1984) Evolution strategies: A family of non-linear optimization techniques based on imitating some principles of organic evolution. Annals of Operations Research 1 (2):165-167
8. Glover F (1986) Future paths for integer programming and links to artificial intelligence. Computers & Operations Research 13 (5):533-549. doi:https://doi.org/10.1016/0305-0548(86)90048-1
9. Van Laarhoven PJ, Aarts EH (1987) Simulated annealing. In:  Simulated annealing: Theory and applications. Springer, pp 7-15
10. Storn R, Price K (1997) Differential Evolution – A Simple and Efficient Heuristic for global Optimization over Continuous Spaces. Journal of Global Optimization 11 (4):341-359. doi:10.1023/A:1008202821328
11. Simon D (2008) Biogeography-Based Optimization. IEEE Transactions on Evolutionary Computation 12 (6):702-713. doi:10.1109/TEVC.2008.919004
12. Ghaemi M, Feizi-Derakhshi M-R (2014) Forest Optimization Algorithm. Expert Systems with Applications 41 (15):6676-6687. doi:https://doi.org/10.1016/j.eswa.2014.05.009

13. Eberhart R, Kennedy J A new optimizer using particle swarm theory. In: MHS'95. Proceedings of the Sixth International Symposium on Micro Machine and Human Science, 4-6 Oct. 1995 1995. pp 39-43. doi:10.1109/MHS.1995.494215
14. Dorigo M, Caro GD Ant colony optimization: a new meta-heuristic. In: Proceedings of the 1999 Congress on Evolutionary Computation-CEC99 (Cat. No. 99TH8406), 6-9 July 1999 1999. pp 1470-1477 Vol. 1472. doi:10.1109/CEC.1999.782657
15. Chu S-C, Tsai P-w, Pan J-S Cat Swarm Optimization. In: Yang Q, Webb G (eds) PRICAI 2006: Trends in Artificial Intelligence, Berlin, Heidelberg, 2006// 2006. Springer Berlin Heidelberg, pp 854-858
16. Karaboga D, Basturk B (2007) A powerful and efficient algorithm for numerical function optimization: artificial bee colony (ABC) algorithm. Journal of Global Optimization 39 (3):459-471. doi:10.1007/s10898-007-9149-x
17. Yang X-S, Deb S Cuckoo search via Lévy flights. In: 2009 World congress on nature & biologically inspired computing (NaBIC), 2009. IEEE, pp 210-214
18. Yang X-S Firefly algorithms for multimodal optimization. In: International symposium on stochastic algorithms, 2009. Springer, pp 169-178
19. Li X, Zhang J, Yin M (2014) Animal migration optimization: an optimization algorithm inspired by animal migration behavior. Neural Computing and Applications 24 (7):1867-1877. doi:10.1007/s00521-013-1433-8
20. Kaveh A, Farhoudi N (2013) A new optimization method: Dolphin echolocation. Advances in Engineering Software 59:53-70. doi:https://doi.org/10.1016/j.advengsoft.2013.03.004
21. Khishe M, Mosavi MR (2020) Chimp optimization algorithm. Expert Systems with Applications 149:113338. doi:https://doi.org/10.1016/j.eswa.2020.113338
22. Mousavirad SJ, Ebrahimpour-Komleh H (2017) Human mental search: a new population-based metaheuristic optimization algorithm. Applied Intelligence 47 (3):850-887. doi:10.1007/s10489-017-0903-6
23. Fausto F, Cuevas E, Valdivia A, González A (2017) A global optimization algorithm inspired in the behavior of selfish herds. Biosystems 160:39-55. doi:https://doi.org/10.1016/j.biosystems.2017.07.010
24. Abualigah L (2020) Group search optimizer: a nature-inspired meta-heuristic optimization algorithm with its results, variants, and applications. Neural Computing and Applications. doi:10.1007/s00521-020-05107-y
25. Tang D, Dong S, Jiang Y, Li H, Huang Y (2015) ITGO: Invasive tumor growth optimization algorithm. Applied Soft Computing 36:670-698. doi:https://doi.org/10.1016/j.asoc.2015.07.045
26. Apolloni B, Carvalho C, de Falco D (1989) Quantum stochastic optimization. Stochastic Processes and their Applications 33 (2):233-244. doi:https://doi.org/10.1016/0304-4149(89)90040-9
27. Formato RA (2007) Central force optimization. Prog Electromagn Res 77:425-491
28. Rashedi E, Nezamabadi-pour H, Saryazdi S (2009) GSA: A Gravitational Search Algorithm. Information Sciences 179 (13):2232-2248. doi:https://doi.org/10.1016/j.ins.2009.03.004
29. Kaveh A, Talatahari S (2010) A novel heuristic optimization method: charged system search. Acta Mechanica 213 (3):267-289. doi:10.1007/s00707-009-0270-4
30. Hatamlou A (2013) Black hole: A new heuristic optimization approach for data clustering. Information Sciences 222:175-184. doi:https://doi.org/10.1016/j.ins.2012.08.023
31. Husseinzadeh Kashan A (2015) A new metaheuristic for optimization: Optics inspired optimization (OIO). Computers & Operations Research 55:99-125. doi:https://doi.org/10.1016/j.cor.2014.10.011
32. Kaveh A, Bakhshpoori T (2016) Water Evaporation Optimization: A novel physically inspired optimization algorithm. Computers & Structures 167:69-85. doi:https://doi.org/10.1016/j.compstruc.2016.01.008

33. Punnathanam V, Kotecha P (2016) Yin-Yang-pair Optimization: A novel lightweight optimization algorithm. Engineering Applications of Artificial Intelligence 54:62-79. doi:https://doi.org/10.1016/j.engappai.2016.04.004
34. Kaveh A, Dadras A (2017) A novel meta-heuristic optimization algorithm: Thermal exchange optimization. Advances in Engineering Software 110:69-84. doi:https://doi.org/10.1016/j.advengsoft.2017.03.014
35. Abedinpourshotorban H, Mariyam Shamsuddin S, Beheshti Z, Jawawi DNA (2016) Electromagnetic field optimization: A physics-inspired metaheuristic optimization algorithm. Swarm and Evolutionary Computation 26:8-22. doi:https://doi.org/10.1016/j.swevo.2015.07.002
36. Kaveh A, Khanzadi M, Rastegar Moghaddam M (2020) Billiards-inspired optimization algorithm; a new meta-heuristic method. Structures 27:1722-1739. doi:https://doi.org/10.1016/j.istruc.2020.07.058
37. Gan C, Cao W, Wu M, Chen X (2018) A new bat algorithm based on iterative local search and stochastic inertia weight. Expert Systems with Applications 104:202-212. doi:https://doi.org/10.1016/j.eswa.2018.03.015
38. Wolpert DH, Macready WG (1997) No free lunch theorems for optimization. IEEE Transactions on Evolutionary Computation 1 (1):67-82. doi:10.1109/4235.585893
39. contributors W (2020) Call of Duty: Warzone — Wikipedia, The Free Encyclopedia.
40. contributors W (2020) Apex Legends — Wikipedia, The Free Encyclopedia.
41. contributors W (2020) Counter-Strike: Global Offensive — Wikipedia, The Free Encyclopedia.
42. contributors W (2020) Ring of Elysium — Wikipedia, The Free Encyclopedia.
43. Mirjalili S (2015) The Ant Lion Optimizer. Advances in Engineering Software 83:80-98. doi:https://doi.org/10.1016/j.advengsoft.2015.01.010
44. Mirjalili S (2015) Moth-flame optimization algorithm: A novel nature-inspired heuristic paradigm. Knowledge-Based Systems 89:228-249. doi:https://doi.org/10.1016/j.knosys.2015.07.006
45. Mirjalili S, Mirjalili SM, Hatamlou A (2016) Multi-Verse Optimizer: a nature-inspired algorithm for global optimization. Neural Computing and Applications 27 (2):495-513. doi:10.1007/s00521-015-1870-7
46. Mirjalili S (2016) SCA: A Sine Cosine Algorithm for solving optimization problems. Knowledge-Based Systems 96:120-133. doi:https://doi.org/10.1016/j.knosys.2015.12.022
47. Mirjalili S, Lewis A (2016) The Whale Optimization Algorithm. Advances in Engineering Software 95:51-67. doi:https://doi.org/10.1016/j.advengsoft.2016.01.008